\documentclass[10pt,twocolumn,letterpaper]{article}

\usepackage{subfiles}
\usepackage{titling} 

\usepackage{titlesec}

\titleformat{\paragraph}[runin]
{\normalfont\normalsize\bfseries}
{}{0pt}{}[:]

\titlespacing*{\paragraph}{0pt}{1\baselineskip}{1em}

\usepackage[algorithms]{wacv}

\usepackage{graphicx}
\usepackage{amsmath}
\usepackage{amssymb}
\usepackage{booktabs}

\usepackage[font=footnotesize]{subcaption}

\usepackage[separate-uncertainty=true, per-mode=symbol]{siunitx}
\usepackage{tabularx}
\usepackage{mathtools}
\usepackage{dsfont}
\usepackage{pifont}

\usepackage{xcolor,colortbl}
\definecolor{Gray}{gray}{0.85}
\newcolumntype{a}{>{\columncolor{Gray}}c}

\usepackage[pagebackref,breaklinks,colorlinks]{hyperref}
\usepackage[capitalize]{cleveref}
\crefname{section}{Sec.}{Secs.}
\Crefname{section}{Section}{Sections}
\Crefname{table}{Table}{Tables}
\crefname{table}{Tab.}{Tabs.}

\begin{document}

\title{Memory-Efficient Pseudo-Labeling for Online Source-Free Universal Domain Adaptation using a Gaussian Mixture Model}

\author{Pascal Schlachter, Simon Wagner, Bin Yang\\
University of Stuttgart, Germany\\
{\tt\small \{pascal.schlachter, bin.yang\}@iss.uni-stuttgart.de}
}
\maketitle

\begin{abstract}
	In practice, domain shifts are likely to occur between training and test data, necessitating domain adaptation (DA) to adjust the pre-trained source model to the target domain. Recently, universal domain adaptation (UniDA) has gained attention for addressing the possibility of an additional category (label) shift between the source and target domain. This means new classes can appear in the target data, some source classes may no longer be present, or both at the same time. For practical applicability, UniDA methods must handle both source-free and online scenarios, enabling adaptation without access to the source data and performing batch-wise updates in parallel with prediction. In an online setting, preserving knowledge across batches is crucial. However, existing methods often require substantial memory, which is impractical because memory is limited and valuable, in particular on embedded systems. Therefore, we consider memory-efficiency as an additional constraint. To achieve memory-efficient online source-free universal domain adaptation (SF-UniDA), we propose a novel method that continuously captures the distribution of known classes in the feature space using a Gaussian mixture model (GMM). This approach, combined with entropy-based out-of-distribution detection, allows for the generation of reliable pseudo-labels. Finally, we combine a contrastive loss with a KL divergence loss to perform the adaptation. Our approach not only achieves state-of-the-art results in all experiments on the DomainNet and Office-Home datasets but also significantly outperforms the existing methods on the challenging VisDA-C dataset, setting a new benchmark for online SF-UniDA. Our code is available at \url{https://github.com/pascalschlachter/GMM}.
\end{abstract}

\section{Introduction}
Despite their success, deep neural networks still struggle when there is a domain (distribution) shift between the training and test data, which is a common problem in real-world applications. To tackle this challenge, unsupervised domain adaptation (UDA) is applied to adapt the pre-trained source model to the unlabeled target data. While classical UDA assumes the availability of the source dataset during adaptation, recent test-time adaptation (TTA) continuously adapts the pre-trained source model to the target domain(s) during deployment of the model using only the unlabeled test-time data. Accordingly, TTA is also known as source-free UDA. On one hand, being independent of the source dataset enables UDA even when the source data is inaccessible due to privacy concerns or memory constraints. On the other hand, by refraining to reprocess the source data during testing the computational efficiency is enhanced \cite{tent}.

\begin{figure}[!t]
	\centering
	\includegraphics[width=0.8\linewidth]{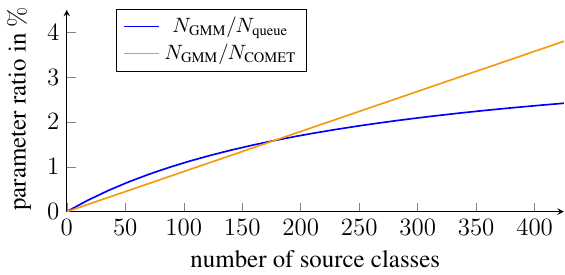}
	\caption{Comparison of the memory required for the knowledge transfer between our GMM-based method, a memory queue \cite{AdaContrast}, and an additional teacher model like used by COMET \cite{comet}. Our GMM-based method is clearly the most memory-efficient.}
	\label{fig:memory}
\end{figure}

However, the vast majority of approaches towards TTA for classification only work in the standard closed-set setting, meaning they assume the source label space $\mathcal{Y}_\mathrm{s}$ to be identical to the target label space $\mathcal{Y}_\mathrm{t}$. This greatly limits the applicability for practical problems which are often open-world scenarios and therefore violate this assumption. To overcome this limitation and handle both a domain and a category (label) shift at the same time, source-free universal domain adaptation (SF-UniDA) \cite{sf_unida} was introduced. Unlike previous approaches tailored to either partial-set DA (PDA), open-set DA (ODA) or open-partial-set DA (OPDA), SF-UniDA considers that usually no prior knowledge of the category shift is available. Thus, SF-UniDA aims to universally manage any kind of category shift.

In this way, SF-UniDA is related to few-shot learning \cite{few_shot} and class-incremental continual learning \cite{class_incremental}, both of which aim to enhance a pre-trained source model's ability to accurately classify new classes using a small labeled target dataset. However, SF-UniDA differs in that it is unsupervised and thus does not rely on labeled target data. Consequently, in open-set or open-partial-set scenarios, SF-UniDA aims only to reject samples of new classes as unknown. Moreover, SF-UniDA addresses a simultaneous domain shift in addition to a category shift.

Besides category shifts, TTA was recently also extended to the online application. This advancement allows to adapt the source model continuously to the target data in parallel to inference, rather than relying on a fixed adaptation performed beforehand. Hence, each test batch is only accessed once and only one batch at a time. This is not only computationally more efficient but especially allows the application of TTA when, instead of a finite test set with unlimited access, real-time predictions for a data stream are required. 

Recently, \cite{comet} showed that most existing approaches towards SF-UniDA \cite{sf_unida, glc_plus, lead} do not generalize well to the online scenario since the individual batches cannot sufficiently represent their underlying data distribution. Therefore, approaches tackling this limitation aim to save the knowledge of previous batches either using a memory queue \cite{AdaContrast} or a mean teacher architecture \cite{comet}. However, both approaches require substantial memory which is a limited and valuable resource in practice.

Accordingly, in this paper, we introduce memory-efficiency as an additional constraint for online SF-UniDA, aiming to create a more realistic scenario and enhance applicability. Consequently, we propose a novel method designed to achieve memory-efficient online SF-UniDA. It uses a Gaussian mixture model (GMM) to adaptively capture the underlying distribution of the known classes in the target data within the feature space with only few parameters. In combination with an entropy-based out-of-distribution (OOD) detection, the GMM provides reliable pseudo-labels in the online scenario while remaining memory-efficient as shown in \cref{fig:memory}. Subsequently, for adaptation, we apply a combination of a contrastive loss and a Kullback-Leibler (KL) divergence loss to get a meaningful feature space and clear predictions.

For evaluation, we test our GMM-based method on the public DA datasets DomainNet, VisDA-C and Office-Home where it achieves state-of-the-art results despite being strictly source-free and memory-efficient. Remarkably, on the VisDA-C dataset, our GMM-based method can even set a new state-of-the-art for online SF-UniDA. This applies in particular to the OPDA scenario, where the margin to the second-best method is more than $\SI{13}{\percent}$.


\section{Related work}
\subsection{Online test-time adaptation}
Methods towards online TTA can be mainly classified into three categories. First, batch normalization (BN) statistics calibration methods \cite{tent, niu2023towards} adapt the BN layers of the source model to the shifted statistics of the target data. Second, self-training methods \cite{AdaContrast, shot, goyal2022test, dobler2023robust} generate pseudo-labels for the target data to perform the adaptation in a supervised way. Third, entropy minimization methods \cite{tent, shot, pmlr-v162-niu22a} adapt the model such that the entropy of its outputs is minimized. However, all methods mentioned above do not consider the possibility of a category shift in addition to the domain shift.

\subsection{Universal domain adaptation}

Although numerous approaches address both domain and category shifts, most of them \cite{cao2018partial, cao2018partial2, zhang2018importance, open_set_domain_adaptation, saito2018open, psdc, liu2019separate, busto2018open, baktashmotlagh2018learning, bucci2020effectiveness, you2019universal, fu2020learning, saito2021ovanet, saito2020universal, Li_2021_CVPR, chen2022geometric, lu2024mlnet} depend on access to source data. Additionally, some methods \cite{universal_sf_da, kundu2020towards, umad, coca} claim to be source-free because they do not reprocess the source data during adaptation, but they require a specialized, e.g. open-set, source model training and/or architecture. This reliance significantly limits their practical applicability. Following the strict definition given by \cite{sf_unida}, we do not consider these methods truly source-free since they depend on more source information than just a standard pre-trained source model. Among the remaining approaches, \cite{shot, kundu2020towards, li2023robustness, feng2021open} are not universal as they are either specialized for a single category shift or require prior knowledge about the type of shift.

The only approaches that are both truly source-free and universal are \cite{sf_unida, glc_plus, lead, comet}, with \cite{comet} being the sole method additionally designed for the online setting. \cite{sf_unida} introduces a clustering-based pseudo-labeling for SF-UniDA called Global and Local Clustering (GLC). For adaptation, they combine a cross-entropy loss with a kNN-based loss. Finally, during inference, they apply an entropy threshold to reject samples as unknown. \cite{glc_plus} extends this approach by a contrastive affinity learning strategy realized by an additional contrastive loss. \cite{lead} also builds on the work of \cite{sf_unida} by using the same kNN-based loss and entropy-based rejection strategy for inference. However, they introduce a different pseudo-labeling approach called Learning Decomposition (LEAD) which divides features into source-known and -unknown components. This allows to build instance-level decision boundaries to identify target-private data during pseudo-labeling.
For adaptation, they apply a confidence-weighted cross-entropy loss and a feature decomposition regularizer besides the kNN-based loss. Finally, \cite{comet} introduces a method called Contrastive Mean Teacher (COMET) specially tailored to online SF-UniDA. It combines a contrastive and an
entropy loss embedded into a mean teacher framework for pseudo-labeling.


\section{Method}
\begin{figure*}[!t]
	\centering
	\includegraphics[width=0.89\linewidth]{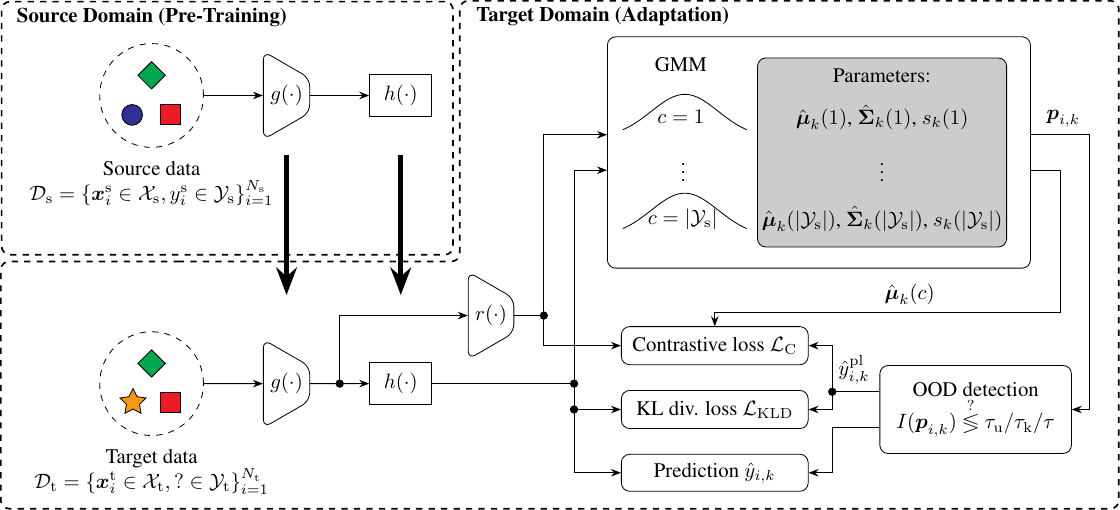}
	\caption{Overview of our proposed GMM-based approach. After pre-training on the source data $\mathcal{D}_\mathrm{s}$, the model is adapted to the target data $\mathcal{D}_\mathrm{t}$ using a combination of the two losses $\mathcal{L}_\mathrm{C}$ and $\mathcal{L}_\mathrm{KLD}$. For these losses, reliable pseudo-labels are provided by a GMM that models the distributions of the $|\mathcal{Y}_\mathrm{s}|$ known classes in a reduced feature space. Its parameters are iteratively updated with each target batch. To reject samples from new classes as unknown during both pseudo-labeling and prediction, we apply an entropy-based OOD detection.}
	\label{fig:overview}
\end{figure*}

\subsection{Preliminaries}
Our goal is to achieve universal DA in an online, source-free and memory-efficient manner. DA describes the task to adapt a neural network $f$, pre-trained on the source dataset $\mathcal{D}_\mathrm{s}=\{\boldsymbol{x}_i^\mathrm{s}\in\mathcal{X}_\mathrm{s}, y_i^\mathrm{s}\in\mathcal{Y}_\mathrm{s}\}_{i=1}^{N_\mathrm{s}}$, to the unlabeled target data $\mathcal{D}_\mathrm{t}=\{\boldsymbol{x}_i^\mathrm{t}\in\mathcal{X}_\mathrm{t}, ?\in\mathcal{Y}_\mathrm{t}\}_{i=1}^{N_\mathrm{t}}$. The model can be divided into a feature extractor $g$ and a classifier $h$, i.e. $f=h\circ g$. In contrast to conventional DA, universal DA considers the case that the target dataset $\mathcal{D}_\mathrm{t}$ experiences a category shift in addition to a domain shift. A category shift denotes the scenario that the target label space $\mathcal{Y}_\mathrm{t}$ deviates from the source label space $\mathcal{Y}_\mathrm{s}$ (i.e. $\mathcal{Y}_\mathrm{t}\neq\mathcal{Y}_\mathrm{s}$). This includes three cases: PDA ($\mathcal{Y}_\mathrm{t}\subset \mathcal{Y}_\mathrm{s}$), ODA ($\mathcal{Y}_\mathrm{s}\subset \mathcal{Y}_\mathrm{t}$) and OPDA ($\mathcal{Y}_\mathrm{s}\cap\mathcal{Y}_\mathrm{t}\neq\emptyset$, $\mathcal{Y}_\mathrm{s}\nsubseteq\mathcal{Y}_\mathrm{t}$, $\mathcal{Y}_\mathrm{s}\nsupseteq\mathcal{Y}_\mathrm{t}$). The goal of universal DA is to universally perform well in all three scenarios without any prior knowledge about the category shift. This is the case if samples of new classes are reliably rejected as unknown while samples of known classes are classified correctly.

The paradigm source-free describes the constraint that the source dataset $\mathcal{D}_\mathrm{s}$ cannot be accessed during the adaptation. Hence, the standard pre-trained source model is the only source knowledge that can be leveraged.  The second paradigm online denotes that the target data is not available as a dataset with unlimited access but instead is received as a stream of batches. This means that at any time we only have the current target batch $\{\boldsymbol{x}_{i, k}^\mathrm{t}\}_{i=1}^{N_\mathrm{b}}$ of size $N_\mathrm{b}$ for adaptation which demands an immediate prediction. Thereby, $k$ represents the batch number. In this scenario, saving knowledge from previously processed batches can be advantageous to support the adaptation. Nevertheless, memory is limited and valuable in practice. Hence, the third paradigm we consider is to save this knowledge as memory-efficiently as possible. By combining all three paradigms our proposed GMM-based approach features maximal applicability for real-world and embedded applications, where the source data is often subject to privacy policies, real-time inference of data streams may be required and hardware constraints must be met. \cref{fig:overview} shows an overview of our method.

\subsection{GMM-based pseudo-labeling}
\label{pseudo_labeling}
While self-training through pseudo-labeling has already proven to be successful for SF-UniDA \cite{sf_unida, glc_plus, lead}, generating reliable pseudo-labels online is challenging due to the limited available data at any time. Clustering and kNN-based methods, in particular, perform poorly because individual target batches do not sufficiently represent the underlying data distribution \cite{comet}.

To address this, \cite{AdaContrast} proposed saving target samples into a memory queue to increase the available amount of data. They show that this can enable a kNN-based pseudo-labeling. However, since their memory queue stores the feature and prediction vectors of $55,388$ samples it massively increases the memory requirement. A similar issue arises with \cite{comet}, which requires a copy of the model to build a mean teacher framework. Depending on its architecture, this can also demand a substantial amount of memory.

While saving knowledge from previous target batches is essential to get reliable pseudo-labels, this must be done in a memory-efficient way to ensure practical applicability. Therefore, we propose a novel method that achieves memory-efficient knowledge transfer resulting in a reliable pseudo-labeling for online SF-UniDA. Our basic idea is to continuously save the data distribution of the target data in the feature space using a GMM. Concretely, each of the $|\mathcal{Y}_\mathrm{s}|$ known classes is modeled by a Gaussian distribution, representing a mode of the GMM. Although, the underlying assumption that each class follows a unimodal Gaussian distribution in the feature space cannot be guaranteed, the contrastive loss introduced in \cref{sec:contrastive} supports this hypothesis by enforcing the class-wise clustering of samples in the feature space.

For each target batch $\{\boldsymbol{x}_{i, k}^\mathrm{t}\}_{i=1}^{N_\mathrm{b}}$ we first update the GMM parameters $\hat{\boldsymbol{\mu}}_{k}(c)$ and $\hat{\boldsymbol{\Sigma}}_{k}(c)$, i.e. the means and covariance matrices of all modes $1\leq c\leq |\mathcal{Y}_\mathrm{s}|$, by a weighted recalculation inspired by the expectation maximization (EM) algorithm. Subsequently, the resulting GMM enables the calculation of the likelihood $p(\boldsymbol{x}_{i,k}^\mathrm{t}| c;\hat{\boldsymbol{\mu}}_{k}(c),\hat{\boldsymbol{\Sigma}}_{k}(c))$ for a given target sample $\boldsymbol{x}_{i,k}^\mathrm{t}$ belonging to class $c$, thus allowing us to generate reliable pseudo-labels.

The EM algorithm is divided into an E-step and an M-step. In the E-step, the weights are calculated. There are two types of weightings to consider. The first weighting determines how much each sample $\boldsymbol{x}_{i,k}^\mathrm{t}$ within a single target batch $\{\boldsymbol{x}_{i, k}^\mathrm{t}\}_{i=1}^{N_\mathrm{b}}$ contributes to the update of each GMM parameter. Instead of deriving these weights from the GMM, we use the model predictions $f(\boldsymbol{x}_{i,k}^\mathrm{t})$. Second, weighting across batches accounts for the unequal representation of classes in different batches. A batch with more samples of a certain class should contribute more to the parameter update of that class than a batch with fewer samples. We achieve this by using the class-wise sum of all softmax classifier outputs of a batch as an indicator of class representation. By iteratively summing these values and comparing them to the current value, we determine how much the current batch should contribute to the overall parameter value. Accordingly, we define
\begin{align}
	s_{k}(c)= \alpha\cdot s_{k-1}(c) + \sum_{i=1}^{N_\mathrm{b}} f_c(\boldsymbol{x}_{i,k}^\mathrm{t})~,
	\label{eq:storage}
\end{align}
which is initialized by $s_0(c)=0$ $\forall c$. Thereby, $f_c()$ denotes the softmax classifier output corresponding to class $c$ and $\alpha\in[0,1]$ is an exponential decay factor that controls the influence of previous batches.

In the M-step, we recursively update both the mean vector $\boldsymbol{\hat{\mu}}_{k}\left(c\right)$ and covariance matrix $\boldsymbol{\hat{\Sigma}}_{k} \left(c\right)$ for each mode $c$ using the samples of the current target batch $\{\boldsymbol{x}_{i, k}^\mathrm{t}\}_{i=1}^{N_\mathrm{b}}$:
\begin{align}
	\hat{\boldsymbol{\mu}}_{k}(c) = \frac{\alpha\cdot s_{k-1}(c) \cdot \hat{\boldsymbol{\mu}}_{k-1} \left(c\right) + \sum\limits_{i=1}^{N_\mathrm{b}} f_c(\boldsymbol{x}_{i,k}^\mathrm{t}) \cdot r_{i,k}}{s_{k}(c)}
	\label{eq:gmmmu}
\end{align}
\begin{align}
	\hat{\boldsymbol{\Sigma}}_{k}(c)= \frac{\left(\splitdfrac{\hspace*{-0.1cm}\alpha\cdot s_{k-1}(c) \cdot \hat{\boldsymbol{\Sigma}}_{k-1}(c)+ \sum\limits_{i=1}^{N_\mathrm{b}} f_c(\boldsymbol{x}_{i,k}^\mathrm{t})}{\cdot \left(r_{i,k}-\hat{\boldsymbol{\mu}}_{k}(c)\right)\left(r_{i,k}-\hat{\boldsymbol{\mu}}_{k}(c)\right)^T}\hspace*{-0.1cm}\right)}{s_{k}(c)}
	\label{eq:gmmcov}
\end{align}
Thereby, $r_{i,k}$ is an abbreviation for $r(g(\boldsymbol{x}_{i,k}^\mathrm{t}))$ where $r()$ denotes a feature dimension reduction function. We found that applying our GMM-based pseudo-labeling in a lower-dimensional feature space than the output of $g()$ not only reduces the memory requirement of the GMM parameters but also improves performance. We realize $r()$ as a single linear layer, which is optimized in parallel to the model $f$.

Despite the iterative calculation, we do not require initialization of $\hat{\boldsymbol{\mu}}_{0}(c)$ and $\hat{\boldsymbol{\Sigma}}_{0}(c)$ since we initialize $s_0(c)=0$ $\forall c$. Nevertheless, if prior knowledge is available, it can be incorporated by setting $\hat{\boldsymbol{\mu}}_{0}(c)$ and $\hat{\boldsymbol{\Sigma}}_{0}(c)$ accordingly and selecting appropriate weights $s_0(c)$.

After updating the GMM parameters, for each sample $\boldsymbol{x}_{i,k}^\mathrm{t}$ in the current target batch $\{\boldsymbol{x}_{i, k}^\mathrm{t}\}_{i=1}^{N_\mathrm{b}}$ we determine the likelihood $p(\boldsymbol{x}_{i,k}^\mathrm{t}| c;\hat{\boldsymbol{\mu}}_{k}(c),\hat{\boldsymbol{\Sigma}}_{k}(c))$ across all known classes $1\leq c \leq |\mathcal{Y}_\mathrm{s}|$. This is done by evaluating each sample $\boldsymbol{x}_{i,k}^\mathrm{t}$ on all $|\mathcal{Y}_\mathrm{s}|$ Gaussian distributions $\mathcal{N}\left(\hat{\boldsymbol{\mu}}_{k}(c), \hat{\boldsymbol{\Sigma}}_{k}(c)\right)$. The final pseudo-labels for the known classes are then assigned based on the class with the maximum likelihood.

Finally, in \cref{eq:gmmmu} and \cref{eq:gmmcov}, note that each sample contributes to all parameters. In this way, predictions of the GMM are also influenced by similar samples, leading to a reduction of noise in the pseudo-labels.

\subsection{Out-of-distribution detection}
Since the GMM only models known classes, a maximum likelihood (ML) estimation alone would classify all samples as belonging to one of these known classes. However, for online SF-UniDA, it is crucial to identify samples from new classes and pseudo-label them accordingly by introducing a new ``unknown'' class $|\mathcal{Y}_\mathrm{s}|+1$. Therefore, we extend the GMM-based ML estimation with an OOD detection mechanism. We accomplish OOD detection by assessing the confidence of a sample belonging to the GMM distribution. Through empirical investigation (shown in the supplementary material), we found that the normalized Shannon entropy of the likelihoods output by the GMM is most effective for this purpose. This entropy measure is defined as
\begin{align}
	I(\boldsymbol{p}_{i,k})=-\frac{1}{\log |\mathcal{Y}_\mathrm{s}|}\cdot\boldsymbol{p}_{i,k}^T\cdot\log \boldsymbol{p}_{i,k}
	\label{eq:entropy}
\end{align}
where $\boldsymbol{p}_{i,k} = \left[ p(\boldsymbol{x}_{i,k}^\mathrm{t} | c; \hat{\boldsymbol{\mu}}_{k}(c), \hat{\boldsymbol{\Sigma}}_{k}(c)) \right]_{c=1}^{|\mathcal{Y}_\mathrm{s}|}$ represents a vector containing the likelihoods indicating the probability of sample $\boldsymbol{x}_{i,k}^\mathrm{t}$ belonging to all classes $1 \leq c \leq |\mathcal{Y}_\mathrm{s}|$. A low entropy value suggests high confidence that the sample $\boldsymbol{x}_{i,k}^\mathrm{t}$ belongs to the GMM distribution, hence to one of the known classes. Conversely, a high entropy indicates low confidence that the sample belongs to this distribution, suggesting it is likely OOD and belongs to a new class.

Recently, \cite{feng2021open} and \cite{comet} proposed excluding samples with uncertain pseudo-labels from the adaptation process since  they argue that incorrect pseudo-labels lead to a negative transfer being more harmful than using fewer samples for adaptation. Hence, instead of using a single entropy threshold to distinguish samples from new classes and known classes, we introduce two thresholds, $\tau_\mathrm{k}$ and $\tau_\mathrm{u}$ ($\tau_\mathrm{k}<\tau_\mathrm{u}$), to filter out samples with uncertain estimates and not assigning them a pseudo-label. The final pseudo-labeling of a target sample $\boldsymbol{x}_{i,k}^\mathrm{t}$ is described as follows:
\begin{align}
	\hat{y}_{i,k}^\mathrm{pl}=\begin{cases}
		\arg\max\limits_{c\in\mathcal{Y}_\mathrm{s}} {\boldsymbol{p}}_{i,k}
		 & I({\boldsymbol{p}}_{i,k})\leq\tau_\mathrm{k}\\
		|\mathcal{Y}_\mathrm{s}|+1 & I({\boldsymbol{p}}_{i,k})\geq\tau_\mathrm{u}\\
		\text{None} & \text{otherwise}
	\end{cases}
\end{align}
To avoid the unintuitive manual selection of the thresholds $\tau_\mathrm{k}$ and $\tau_\mathrm{u}$, we adaptively initialize them on the fly within the first $N_\mathrm{init}$ batches. Thereby, for each of these batches, we sort the samples w.r.t. $I({\boldsymbol{p}}_{i,k})$ and identify the entropy values that separate the top and bottom $(100-p_\mathrm{reject})/2$ percent of samples. We then calculate $\tau_\mathrm{k}$ and $\tau_\mathrm{u}$ by averaging these entropy values across the previously processed batches. After this initialization period, the thresholds remain fixed for the rest of the adaptation process. In this way, only $p_\mathrm{reject}$ and $N_\mathrm{init}$ need to be chosen manually, which are both well-interpretable for humans in contrast to $\tau_\mathrm{k}$ and $\tau_\mathrm{u}$.

\subsection{Contrastive loss}
\label{sec:contrastive}
Using the pseudo-labels, we apply a contrastive loss \cite{khosla2020supervised} to perform the domain adaptation. It is able to enforce desired properties on the feature space by minimizing the distances of positive pairs of samples and maximizing the distances of negative pairs. \cite{li2023robustness} and \cite{comet} showed that it can be successfully applied for online SF-UniDA. Following their idea, we want to enforce the samples of known classes to build dense and clearly separated clusters in the feature space. Moreover, the samples of new classes should be well-separated from these clusters. 
The contrastive loss achieves this by treating the combinations of each sample of a known class with both the other samples of the same class and the mean of the corresponding GMM mode as positive pairs. Regarding the negative pairs, for each sample of a known class, we maximize the distances to all samples of the different classes, including the ``unknown'' class. Moreover, we maximize the distances between each sample and the GMM's means of all other classes.

Thereby, we extend each batch by adding one augmented version of each sample which not only doubles the number of samples used for adaptation but also leads to a consistent feature space being robust to small changes in the input.

Mathematically, the contrastive loss for the $k$-th target batch $\{\boldsymbol{x}_{i, k}^\mathrm{t}\}_{i=1}^{N_\mathrm{b}}$ is given as
\begin{equation}
	\begin{aligned}
		\mathcal{L}_{\mathrm{C}}\hspace*{-0.05cm} =\hspace*{-0.05cm} &- \hspace*{-0.04cm} \sum_{j=1}^{2N_\mathrm{b}} \sum_{i=1}^{2N_\mathrm{b}} \mathds{1}(\hat{y}_{j,k}^{\mathrm{pl}}\hspace*{-0.07cm}=\hspace*{-0.05cm}\hat{y}_{i,k}^\mathrm{pl}) \log \frac{\exp\hspace*{-0.07cm}\left(\hspace*{-0.04cm}\frac{\langle \tilde{r}_{j,k},\tilde{r}_{i,k}\rangle}{\tau}\hspace*{-0.04cm}\right)}{\sum\limits_{l=1}^{2N_\mathrm{b}}\hspace*{-0.05cm}\exp \hspace*{-0.07cm}\left(\hspace*{-0.04cm}\frac{\langle \tilde{r}_{l,k},\tilde{r}_{i,k}\rangle}{\tau}\hspace*{-0.04cm}\right)} \\
		&-\hspace*{-0.04cm} \sum_{c=1}^{|\mathcal{Y}_\mathrm{s}|} \sum_{i=1}^{2N_\mathrm{b}} \mathds{1}(\hat{y}_{i,k}^\mathrm{pl}\hspace*{-0.07cm}=\hspace*{-0.05cm}c) \log \frac{\exp\hspace*{-0.07cm}\left(\hspace*{-0.04cm}\frac{\langle\hat{\boldsymbol{\mu}}_{k}\hspace*{-0.05cm}(c),\tilde{r}_{i,k}\rangle}{\tau}\hspace*{-0.04cm}\right)}{\sum\limits_{l=1}^{2N_\mathrm{b}}\hspace*{-0.05cm}\exp\hspace*{-0.07cm} \left(\hspace*{-0.04cm}\frac{\langle\hat{\boldsymbol{\mu}}_{k}\hspace*{-0.05cm}(c),\tilde{r}_{l,k}\rangle}{\tau}\hspace*{-0.04cm}\right)}	
		\label{eq:nll}
	\end{aligned}
\end{equation}
where $\mathds{1}(\cdot)$ denotes the indicator function, $\langle\cdot\rangle$ represents the cosine similarity, and $\tau$ is the temperature. Moreover, $\tilde{\boldsymbol{r}}_{i,k}$ is an abbreviation for $r(g(\tilde{\boldsymbol{x}}_{i,k}^\mathrm{t}))$, whereby $\{\tilde{\boldsymbol{x}}_{i,k}^\mathrm{t}\}_{i=1}^{2N_\mathrm{b}}=\{\boldsymbol{x}_{i,k}^\mathrm{t}\}_{i=1}^{N_\mathrm{b}}\cup\{\bar{\boldsymbol{x}}_{i,k}^\mathrm{t}\}_{i=1}^{N_\mathrm{b}}$ denotes the union of the target batch $\{\boldsymbol{x}_{i, k}^\mathrm{t}\}_{i=1}^{N_\mathrm{b}}$ and its augmentation $\{\bar{\boldsymbol{x}}_{i,k}^\mathrm{t}\}_{i=1}^{N_\mathrm{b}}$.

Note that OWTTT \cite{li2023robustness} and COMET-P \cite{comet} use source prototypes, i.e. the class-wise means of the feature representations of the source data, to serve as cluster centers for the contrastive loss. However, the availability of such source prototypes cannot be guaranteed in a source-free setting. In contrast, in our proposed GMM-based method, the cluster centers are given by the means of the individual modes of the GMM. Accordingly, no source prototypes are required which is a significant advantage.

\subsection{KL divergence loss}
While the contrastive loss only optimizes the feature extractor $g$ to get the desired feature space, we combine it with a second loss to additionally optimize the translation between features and predictions performed by the classifier $h$. Thereby, it especially must be taught how to deal with samples of new classes, i.e. feature representations outside of the clusters of known classes, because this is not part of the closed-set source training. Concretely, to support our entropy-based OOD-detection, the goal is to enforce a small entropy for samples of the known classes and a large entropy for samples of new classes. We achieve this using the following KL divergence loss:
\begin{equation}
	\begin{aligned}
	\mathcal{L}_{\mathrm{KLD}} = &-\sum_{i=1}^{N_\mathrm{b}}  \mathds{1}\left(\hat{y}_{i,k}^\mathrm{pl}\in |\mathcal{Y}_\mathrm{s}|\right) D_{\mathrm{KL}}\left(\boldsymbol{u}||f(\boldsymbol{x}_{i,k}^\mathrm{t})\right) \\
	&+ \sum_{i=1}^{N_\mathrm{b}} \mathds{1}\left(\hat{y}_{i,k}^\mathrm{pl}\notin |\mathcal{Y}_\mathrm{s}|\right) D_{\mathrm{KL}}\left(\boldsymbol{u}||f(\boldsymbol{x}_{i,k}^\mathrm{t})\right)
	\label{eq:kldloss}
	\end{aligned}
\end{equation}
where
\begin{equation}
	D_{\mathrm{KL}}\left(p||q\right) = -\sum_ip_i\log q_i + \sum_ip_i\log p_i
	\label{eq:kld}
\end{equation}
denotes the KL divergence between two distributions $p$ and $q$. $\boldsymbol{u}=\left[1/|\mathcal{Y}_\mathrm{s}|,\ldots, 1/|\mathcal{Y}_\mathrm{s}|\right]^T\in\mathbb{R}^{|\mathcal{Y}_\mathrm{s}|}$ serves as a uniform reference distribution. Hence, the loss maximizes the KL divergence between the classifier output and the uniform distribution for samples pseudo-labeled as a known class and minimizes it for samples pseudo-labeled as unknown.

Finally, the overall loss is a combination of the contrastive and KL divergence loss, weighted by $\lambda>0$:
\begin{align}
	\mathcal{L} = \mathcal{L}_\mathrm{C} + \lambda\mathcal{L}_{\mathrm{KLD}}
	\label{eq:loss}
\end{align}

\subsection{Inference}
For inference, we combine standard prediction based on the maximum classifier output with the same OOD detection method used for pseudo-labeling. The only difference is that we now use a single threshold, $\tau$, because every sample must receive a prediction. We set $\tau$ as the average of the two thresholds from pseudo-labeling: $\tau=\frac{\tau_\mathrm{k}+\tau_\mathrm{u}}{2}$. The resulting prediction process for any sample $\boldsymbol{x}_{i,k}^\mathrm{t}$ is as follows:
\begin{align}
	\label{eq:inference}
	\hat{y}_{i,k} = \begin{cases}
		\arg\max f(\boldsymbol{x}_{i,k}^\mathrm{t}) & I(\boldsymbol{p}_{i,k})\leq \tau\\
		|\mathcal{Y}_\mathrm{s}|+1 & I(\boldsymbol{p}_{i,k})> \tau\end{cases}
\end{align}

\subsection{Memory footprint}
\label{sec:memory}
We compare the memory required for the knowledge transfer across batches of our GMM-based method with a memory queue \cite{AdaContrast} and a mean teacher approach \cite{comet}. For each mode, the GMM stores a mean vector $\hat{\boldsymbol{\mu}}_{k}(c)$, a covariance matrix $\hat{\boldsymbol{\Sigma}}_{k}(c)$, and a weight $s_k(c)$. The sizes of $\hat{\boldsymbol{\mu}}_{k}(c)$ and $\hat{\boldsymbol{\Sigma}}_{k}(c)$ depend on the number of dimensions $FD_\mathrm{r}$ in the reduced feature space. Due to its symmetry, $\hat{\boldsymbol{\Sigma}}_{k}(c)$ is fully defined by its upper or lower triangle, including the diagonal. As a result, the total number of parameters is
\begin{align}
	N_\mathrm{GMM}=\left(FD_\mathrm{r} + \frac{FD_\mathrm{r}\cdot(FD_\mathrm{r}+1)}{2} + 1\right)\cdot |\mathcal{Y}_\mathrm{s}|
\end{align}
The memory queue proposed by \cite{AdaContrast} saves the feature and prediction vectors of samples. Thus, in total it stores
\begin{align}
	N_\mathrm{queue}=L\cdot(FD+|\mathcal{Y}_\mathrm{s}|)
\end{align}
values, where $L$ denotes the length of the queue, chosen to be $L=55,388$ in \cite{AdaContrast}, and $FD$ is the number of dimensions in the original feature space.

Using a mean teacher approach like COMET \cite{comet} requires maintaining a copy of the model, so the memory consumption depends on the model architecture. The ResNet-50-based architecture used in the following experiments has approximately $N_\mathrm{COMET}\approx24,000,000$ parameters.

For a fair comparison of the three approaches, we use the same hyperparameters as in the subsequent experiments, i.e. $FD=256$ and $FD_\mathrm{r}=64$. \cref{fig:memory} shows the resulting plots of the ratios $N_\text{GMM}/N_\text{queue}$ and $N_\text{GMM}/N_\text{COMET}$ against the number of source classes $|\mathcal{Y}_\mathrm{s}|$. It is evident that our GMM-based method requires only a small fraction of the memory compared to both other approaches. Specifically, even for $|\mathcal{Y}_\mathrm{s}|=345$ source classes (PDA on DomainNet), our approach requires only approximately $\SI{2.2}{\percent}$ of the memory needed by the memory queue and $\SI{3.1}{\percent}$ compared to the mean teacher. This substantial reduction in memory usage makes the GMM-based approach better suited for embedded applications where memory resources are limited. Further insights are given in the supplementary material.

\section{Experiments}
\begin{table}
	\caption{Class splits, i.e. number of the shared, source-private and target-private classes, for OPDA, ODA and PDA, respectively}
	\label{tab:class_splits}
	\begin{tabularx}{\linewidth}{l*{3}{c}}
		\toprule
		& \multicolumn{3}{c}{$|\mathcal{Y}_\mathrm{s}\cap\mathcal{Y}_\mathrm{t}|$, $|\mathcal{Y}_\mathrm{s}\backslash\mathcal{Y}_\mathrm{t}|$, $|\mathcal{Y}_\mathrm{t}\backslash\mathcal{Y}_\mathrm{s}|$}\\
		\cmidrule(lr){2-4}
		& PDA & ODA & OPDA\\
		\midrule
		DomainNet & 200, 145, 0 & 200, 0, 145 & 150, 50, 145 \\
		VisDA-C & 6, 6, 0 & 6, 0, 6 & 6, 3, 3 \\
		Office-Home & 25, 40, 0 & 25, 0, 40 & 10, 5, 50 \\
		\bottomrule
	\end{tabularx}
\end{table}

\begin{table*}
	\caption{Results of the experiments on DomainNet and VisDA-C. Best results are in red, second best in blue.}
	\label{tab:results}
	\vspace*{-0mm}
	\begin{subtable}[t]{\textwidth}
		\caption{Accuracies in $\%$ for the PDA scenario}
		\label{tab:results_PDA}
		\begin{tabularx}{\textwidth}{p{2.2cm} *{12}{c} a|*{1}{c}}
			\toprule
			PDA & C2P & C2R & C2S & P2C & P2R & P2S & R2C & R2P & R2S & S2C & S2P & S2R & Avg. & V\\
			\midrule
			Source-only & $17.5$ & $31.3$ & $21.0$ & $25.0$ & $41.5$ & $24.0$ & $34.1$ & $30.5$ & $22.7$ & $24.7$ & $17.8$ & $25.2$ & $26.28$ & $17.1$ \\
			\midrule
			
			OWTTT \cite{li2023robustness} & $22.6$ & $28.9$ & $25.1$ & $26.3$ & $28.3$ & $24.6$ & $27.2$ & $26.6$ & $22.9$ & $27.6$ & $19.9$ & $14.5$ & $24.54$ & $28.1$\\
			COMET-P \cite{comet} & $\color{blue}24.8$ & $\color{red}41.2$ & $\color{blue}30.3$ & $\color{blue}34.5$ & $\color{red}51.3$ & $\color{red}35.8$ & $\color{blue}40.6$ & $36.1$ & $30.6$ & $\color{blue}37.5$ & $\color{blue}30.6$ & $\color{red}41.1$ & $\color{blue}36.20$ & $32.1$\\
			SHOT-P \cite{shot} & $9.0$ & $37.1$ & $1.9$ & $0.8$ & $6.4$ & $0.3$ & $1.4$ & $1.0$ & $1.2$ & $0.8$ & $0.2$ & $0.2$ & $5.03$ & $\color{blue}35.6$\\
			\midrule
			GLC \cite{sf_unida} & $12.9$ & $13.4$ & $16.9$ & $18.1$ & $12.1$ & $16.7$ & $16.8$ & $11.0$ & $13.4$ & $23.4$ & $16.5$ & $12.3$ & $15.29$ & $8.2$\\
			GLC++ \cite{glc_plus} & $13.0$ & $13.5$ & $16.7$ & $18.4$ & $13.5$ & $16.9$ & $17.6$ & $11.7$ & $13.5$ & $23.6$ & $16.5$ & $13.3$ & $15.68$ & $10.3$\\
			LEAD \cite{lead} & $6.8$ & $6.2$ & $10.5$ & $13.4$ & $6.9$ & $10.5$ & $9.2$ & $5.5$ & $4.9$ & $18.2$ & $9.0$ & $5.9$ & $8.92$ & $14.4$\\
			COMET-F \cite{comet} & $23.4$ & $38.8$ & $28.3$ & $32.2$ & $\color{blue}48.9$ & $33.7$ & $40.4$ & $\color{blue}36.5$ & $\color{blue}30.7$ & $34.8$ & $28.0$ & $38.3$ & $34.50$ & $31.7$\\
			GMM (Ours) & $\color{red}28.3$ & $\color{blue}41.0$ & $\color{red}32.7$ & $\color{red}34.7$ & $44.6$ & $\color{blue}34.3$ & $\color{red}43.4$ & $\color{red}40.1$ & $\color{red}33.4$ & $\color{red}41.4$ & $\color{red}34.5$ & $\color{blue}40.9$ & $\color{red}37.44$ & $\color{red}41.2$\\
			\bottomrule
		\end{tabularx}
	\end{subtable}
	\vspace*{0mm}
	
	\begin{subtable}[t]{\textwidth}
		\caption{H-scores in $\%$ for the ODA scenario}
		\label{tab:results_ODA}
		\begin{tabularx}{\textwidth}{p{2.2cm} *{12}{c} a|*{1}{c}}
			\toprule
			ODA & C2P & C2R & C2S & P2C & P2R & P2S & R2C & R2P & R2S & S2C & S2P & S2R & Avg. & V\\
			\midrule
			Source-only & $36.3$ & $51.3$ & $39.9$ & $41.2$ & $\color{blue}57.8$ & $40.4$ & $50.2$ & $47.0$ & $38.4$ & $44.4$ & $30.7$ & $45.9$ & $43.63$ & $31.7$\\
			\midrule
			OWTTT \cite{li2023robustness} & $42.1$ & $41.9$ & $45.0$ & $45.3$ & $47.0$ & $42.9$ & $48.7$ & $35.9$ & $40.8$ & $51.9$ & $\color{blue}45.4$ & $44.1$ & $44.25$ & $\color{blue}56.7$\\
			COMET-P \cite{comet} & $\color{blue}42.4$ & $\color{red}54.8$ & $\color{red}46.8$ & $\color{red}47.9$ & $57.2$ & $\color{red}47.7$ & $\color{red}53.8$ & $\color{red}50.8$ & $\color{red}46.3$ & $\color{red}52.7$ & $39.0$ & $\color{blue}52.7$ & $\color{red}49.34$ & $50.9$\\
			SHOT-O \cite{shot} & $41.6$ & $53.9$ & $45.2$ & $46.3$ & $\color{red}59.1$ & $45.2$ & $52.6$ & $49.9$ & $43.6$ & $50.1$ & $33.7$ & $50.4$ & $47.63$ & $54.3$\\
			\midrule
			GLC \cite{sf_unida} & $30.6$ & $34.4$ & $35.8$ & $32.4$ & $26.2$ & $30.3$ & $30.1$ & $23.8$ & $22.5$ & $43.1$ & $32.1$ & $30.1$ & $30.95$ & $8.6$\\
			GLC++ \cite{glc_plus} & $30.2$ & $33.7$ & $35.6$ & $32.1$ & $27.3$ & $29.8$ & $29.5$ & $22.8$ & $21.5$ & $42.0$ & $31.3$ & $30.8$ & $30.55$ & $9.3$\\
			LEAD \cite{lead} & $30.9$ & $38.6$ & $36.5$ & $34.0$ & $36.6$ & $30.8$ & $33.3$ & $27.6$ & $23.2$ & $44.5$ & $33.7$ & $36.3$ & $33.83$ & $19.7$\\
			COMET-F \cite{comet} & $40.7$ & $53.2$ & $45.0$ & $46.2$ & $57.2$ & $\color{blue}46.2$ & $\color{blue}53.0$ & $\color{blue}50.3$ & $\color{blue}45.3$ & $50.0$ & $37.0$ & $51.3$ & $47.95$ & $49.0$\\
			GMM (Ours) & $\color{red}43.3$ & $\color{blue}54.4$ & $\color{blue}46.7$ & $\color{blue}47.0$ & $56.2$ & $44.0$ & $50.7$ & $49.9$ & $44.2$ & $\color{blue}52.6$ & $\color{red}45.7$ & $\color{red}54.9$ & $\color{blue}49.13$ & $\color{red}59.9$\\
			\bottomrule
		\end{tabularx}
		
	\end{subtable}
	\vspace*{0mm}
	
	\begin{subtable}[t]{\textwidth}
		\caption{H-scores in $\%$ for the OPDA scenario}
		\label{tab:results_OPDA}
		\begin{tabularx}{\textwidth}{p{2.2cm} *{12}{c} a|*{1}{c}}
			\toprule
			OPDA & C2P & C2R & C2S & P2C & P2R & P2S & R2C & R2P & R2S & S2C & S2P & S2R & Avg. & V\\
			\midrule
			Source-only & $39.5$ & $55.4$ & $42.7$ & $42.0$ & $57.6$ & $38.8$ & $51.7$ & $47.6$ & $38.6$ & $46.8$ & $32.2$ & $47.9$ & $45.07$ & $26.9$\\
			\midrule
			OWTTT \cite{li2023robustness} & $45.3$ & $49.5$ & $46.9$ & $47.0$ & $45.9$ & $43.9$ & $49.3$ & $40.5$ & $41.0$ & $52.3$ & $\color{blue}46.4$ & $36.6$ & $45.38$ & $\color{blue}46.8$\\
			COMET-P \cite{comet} & $\color{blue}45.6$ & $\color{red}58.6$ & $\color{red}48.9$ & $\color{blue}47.8$ & $57.7$ & $\color{red}46.8$ & $\color{red}55.5$ & $\color{red}51.5$ & $\color{red}46.2$ & $\color{red}54.1$ & $39.0$ & $\color{blue}54.2$ & $\color{blue}50.49$ & $42.9$\\
			SHOT-O \cite{shot} & $44.8$ & $\color{blue}57.9$ & $47.8$ & $47.2$ & $\color{red}59.3$ & $43.7$ & $54.0$ & $50.1$ & $42.8$ & $52.0$ & $35.3$ & $52.3$ & $48.93$ & $45.7$\\
			\midrule
			GLC \cite{sf_unida} & $33.8$ & $40.5$ & $38.6$ & $35.7$ & $32.0$ & $33.1$ & $34.2$ & $28.9$ & $26.0$ & $45.7$ & $35.1$ & $36.1$ & $34.98$ & $11.2$\\
			GLC++ \cite{glc_plus} & $33.4$ & $39.9$ & $38.3$ & $35.0$ & $33.5$ & $32.6$ & $33.6$ & $28.1$ & $25.7$ & $45.1$ & $34.7$ & $35.7$ & $34.63$ & $14.1$\\
			LEAD \cite{lead} & $33.3$ & $42.0$ & $37.4$ & $35.0$ & $37.4$ & $31.1$ & $33.5$ & $30.1$ & $25.2$ & $45.6$ & $34.9$ & $37.9$ & $35.28$ & $23.8$\\
			COMET-F \cite{comet} & $44.1$ & $57.1$ & $47.0$ & $47.7$ & $\color{blue}57.9$ & $\color{blue}45.5$ & $\color{blue}54.8$ & $\color{blue}51.3$ & $\color{blue}45.4$ & $51.7$ & $37.4$ & $53.1$ & $49.42$ & $42.0$\\
			GMM (Ours) & $\color{red}46.4$ & $57.6$ & $\color{blue}48.4$ & $\color{red}48.5$ & $57.1$ & $44.2$ & $51.9$ & $50.9$ & $44.8$ & $\color{blue}53.9$ & $\color{red}46.7$ & $\color{red}56.3$ & $\color{red}50.56$ & $\color{red}60.3$\\
			\bottomrule
		\end{tabularx}
		
	\end{subtable}
	
\end{table*}
\subsection{Setup}
\paragraph{Datasets}
To evaluate our GMM-based method, we utilize the public DA datasets DomainNet \cite{domainnet}, VisDA-C \cite{visda} and Office-Home \cite{office}. From the DomainNet dataset, we use four domains: clipart (C), painting (P), real (R), and sketch (S). Each domain contains between approx. 48,000 and 173,000 images, distributed across 345 classes. The VisDA-C (V) dataset consists of 12 classes and only one domain shift from renderings of 3D models to real-world images from the Microsoft COCO dataset \cite{Coco}. As intended, we use the first as the source domain having 152,397 images and the latter as target domain with 55,388 images. Finally, the Office-Home dataset includes 15,500 images divided into four domains: art (Ar), clipart (Cl), product (Pr), and real world (Re), with each domain containing 65 classes. Since the individual domains are too small to effectively evaluate online adaptation, we mix the three non-source domains to get sufficiently large target datasets.

To construct the category shifts, we take the first $|\mathcal{Y}_\mathrm{s}|$ of the alphabetically ordered classes for source training and the last $|\mathcal{Y}_\mathrm{t}|$ for the target domain. Obviously, the overlapping $|\mathcal{Y}_\mathrm{s}\cap\mathcal{Y}_\mathrm{t}|$ classes are shared between both domains. The class splits we used to create the three category shifts for each of the two datasets are given in \cref{tab:class_splits}.

\begin{table*}
	\caption{Accuracies (for PDA) and H-scores (for ODA and OPDA) in $\%$ for Office-Home. Best results are in red, second best in blue.
	}
	\label{tab:results2}
	\vspace*{-0mm}
	\begin{tabularx}{\textwidth}{p{2.7cm} *{14}{c}}
		\toprule
		& \multicolumn{4}{c}{PDA} & & \multicolumn{4}{c}{ODA} & & \multicolumn{4}{c}{OPDA} \\
		\cmidrule(lr){2-5} \cmidrule(lr){7-10} \cmidrule(lr){12-15}
		Source domain & Ar & Cl & Pr & Re & \hspace{0.4cm} & Ar & Cl & Pr & Re & \hspace{0.4cm} & Ar & Cl & Pr & Re\\
		\midrule
		Source-only & $44.4$ & $33.3$ & $32.8$ & $42.3$ && $62.2$ & $54.2$ & $52.3$ & $59.9$ && $69.9$ & $61.4$ & $60.2$ & $61.4$\\
		\midrule
		OWTTT \cite{li2023robustness} & $34.0$ & $28.0$ & $29.8$ & $32.2$ && $58.9$ & $\color{red}61.0$ & $\color{blue}57.2$ & $57.4$ && $69.2$ & $\color{red}71.8$ & $64.2$ & $65.3$\\
		COMET-P \cite{comet} & $\color{blue}49.9$ & $43.2$ & $36.4$ & $\color{blue}47.1$ && $\color{blue}62.2$ & $58.4$ & $55.5$ & $59.4$ && $\color{blue}70.2$ & $65.3$ & $\color{blue}66.0$ & $\color{red}66.4$\\
		SHOT-O/P \cite{shot} & $\color{red}50.0$ & $\color{red}48.4$ & $\color{blue}41.7$ & $43.7$ && $\color{red}63.3$ & $57.5$ & $57.1$ & $\color{red}61.3$ && $\color{red}70.4$ & $65.3$ & $\color{blue}66.0$ & $65.1$\\
		\midrule
		GLC \cite{sf_unida} & $21.1$ & $20.3$ & $18.0$ & $20.2$ && $37.0$ & $40.4$ & $37.8$ & $38.2$ && $51.1$ & $52.6$ & $50.5$ & $52.3$\\
		GLC++ \cite{glc_plus} & $21.1$ & $22.4$ & $20.3$ & $22.3$ && $35.5$ & $35.9$ & $35.5$ & $37.1$ && $49.5$ & $50.0$ & $50.0$ & $52.9$\\
		LEAD \cite{lead} & $17.6$ & $15.5$ & $13.7$ & $18.4$ && $25.8$ & $25.0$ & $25.1$ & $28.4$ && $23.5$ & $22.6$ & $22.0$ & $25.9$\\
		COMET-F \cite{comet} & $49.1$ & $40.7$ & $36.1$ & $45.4$ && $61.8$ & $57.5$ & $55.1$ & $58.9$ && $69.7$ & $64.3$ & $64.7$ & $65.0$\\
		GMM (Ours) & $46.6$ & $\color{blue}46.5$ & $\color{red}43.2$ & $\color{red}48.3$ && $60.5$ & $\color{blue}60.2$ & $\color{red}58.7$ & $\color{blue}60.1$ && $68.8$ & $\color{blue}65.4$ & $\color{red}66.6$ & $\color{blue}65.8$\\
		\bottomrule
	\end{tabularx}
	
\end{table*}

\paragraph{Competing methods}
Naturally, using source information or performing offline adaptation yields better results. Therefore, we evaluate all methods in the same source-free and online setting to ensure a fair comparison.

First, as a baseline, we apply the source model without adaptation, using an entropy threshold to reject unknown samples. Second, we use GLC \cite{sf_unida}, GLC++ \cite{glc_plus} and LEAD \cite{lead}, which are designed for offline SF-UniDA but are applied batch-wise in the online scenario. Third, we evaluate the modified versions of SHOT \cite{shot}, SHOT-O for OPDA and ODA and SHOT-P for PDA, proposed for category shift scenarios. Fourth, we use OWTTT \cite{li2023robustness}, which, although initially developed for strong OOD detection in online TTA, is also applicable for online SF-UniDA as shown by \cite{comet}. Finally, we apply both versions of COMET \cite{comet}, COMET-P and COMET-F, the only methods specifically designed for online SF-UniDA.

Note that OWTTT \cite{li2023robustness} and COMET-P \cite{comet} use source prototypes which may not be considered source-free and limits their real-world applicability. Additionally, SHOT-O and SHOT-P are not universal, as prior knowledge of the category shift is needed to select the appropriate version.

\paragraph{Implementation details}
To ensure a fair comparison, we use the same pre-trained source model for all methods. It is based on a ResNet-50 \cite{he2016deep} architecture with a $256$-dimensional feature space trained in a similar way like for \cite{shot}, \cite{sf_unida}, and \cite{comet}. For adaptation, we apply a SGD optimizer with momentum $0.9$ and a learning rate of $0.01$ for VisDA-C and $0.001$ for DomainNet and Office-Home. Moreover, we choose the batch size to $N_\mathrm{b}=64$ for DomainNet and VisDA-C and $N_\mathrm{b}=32$ for Office-Home. Regarding the hyperparameters of our GMM-based approach, we decide to reduce the number of features to $FD_\mathrm{r}=64$ and use $\alpha=0.999$ as the exponential decay factor. Furthermore, regarding the OOD detection, we choose the number of batches used for initialization of $\tau_\mathrm{k}$ and $\tau_\mathrm{u}$ to $N_\mathrm{init}=30$ and the ratio of left out samples to $p_\mathrm{reject}=\SI{50}{\percent}$ for DomainNet and VisDA-C and $p_\mathrm{reject}=\SI{25}{\percent}$ for Office-Home. Finally, we select the temperature for the contrastive loss to $\tau=0.1$ and use $\lambda=1$ to weigh both losses equally. Thorough ablation studies, including analyses of hyperparameter choices, are provided in the supplementary material.

As customary, we report the H-score (harmonic mean of known and new class accuracies) for ODA and OPDA, and standard classification accuracy for PDA. Each experiment is run six times, and we report the mean.

\subsection{Results}
The results are shown in \cref{tab:results,tab:results2}. As expected, GLC, GLC++ and LEAD perform poorly in the online setting due to their reliance on clustering and/or kNN. While OWTTT generally shows a solid performance in the ODA and OPDA scenarios, its results in the PDA scenario are less favorable, often falling below the source-only baseline. It is important to note that, although PDA may seem straightforward, it is challenging in the context of UniDA, as it is crucial to minimize the number of samples incorrectly rejected as unknown. SHOT-O and SHOT-P generally deliver strong results, particularly on the Office-Home dataset. However, SHOT-P completely fails on DomainNet in 11 out of the 12 domain shifts, raising concerns about its robustness. We suspect that due to the absence of the diversity promoting loss term in SHOT-P, the entropy minimization may yield the trivial solution in these scenarios.

Although also both versions of COMET consistently perform well, our GMM-based approach appears to have the edge and is overall the most robust method across the different datasets and category shifts. On average, it performs best on DomainNet for OPDA and PDA, and is a close second to COMET-P for ODA, with a margin of just $\SI{0.21}{\percent}$. On Office-Home it always ranks among the top two methods, except when ``art'' is the source domain. Most notably, on the challenging VisDA-C dataset, it is able to clearly outperform all competing methods in all three category shifts, especially for OPDA where the margin to the second best method, OWTTT, is more than $\SI{13}{\percent}$. Remarkably, these results are achieved while being strictly source-free, unlike COMET-P, and requiring significantly less memory than both COMET versions as discussed in \cref{sec:memory} and shown in \cref{fig:memory}.



\section{Conclusion}
Despite its practical importance, we identified memory-efficiency as a so-far overlooked aspect in online SF-UniDA. To address this, we proposed a novel approach that continuously captures the underlying distribution of the target data using a GMM. This enables a memory-efficient knowledge transfer across batches in the online scenario and allows for reliable pseudo-labeling via ML classification together with entropy-based OOD detection. Finally, adaptation is achieved by combining a contrastive and a KL divergence loss. Across the experiments, our proposed method achieved or even improved state-of-the-art results while being strictly source-free and highly memory-efficient.

{\small
\bibliographystyle{ieee_fullname}
\bibliography{refs}
}

	\renewcommand{\thesection}{\Alph{section}}
	\setcounter{section}{0}

\title{Supplementary Material to ``Memory-Efficient Pseudo-Labeling for Online Source-Free Universal Domain Adaptation using a Gaussian Mixture Model''}

\author{Pascal Schlachter, Simon Wagner, Bin Yang\\
	University of Stuttgart, Germany\\
	{\tt\small \{pascal.schlachter, bin.yang\}@iss.uni-stuttgart.de}
}

\maketitle

\thispagestyle{empty}

\section{Ablation studies}
To gain deeper insights into our proposed GMM-based method, we conduct rich ablation studies in the following. Thereby, we use the OPDA scenario on the VisDA-C dataset as a representative example. The OPDA scenario effectively illustrates the challenging trade-off between rejecting new classes and reliably classifying a subset of known classes, while the VisDA-C dataset was selected arbitrarily.

\subsection{Loss functions}
\begin{figure}[!t]
	\centering
	\includegraphics[width=0.9\linewidth]{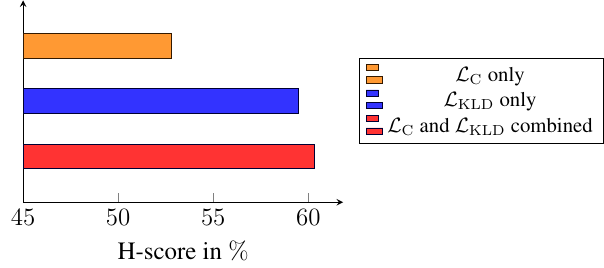}
	\caption{Results using different combinations of the losses $\mathcal{L}_\mathrm{C}$ and $\mathcal{L}_\mathrm{KLD}$ for the VisDA-C OPDA scenario.}
	\label{fig:loss}
\end{figure}

\paragraph{Contribution of each loss}
\cref{fig:loss} illustrates the individual contributions of the contrastive loss $\mathcal{L}_\mathrm{C}$ and the KL divergence loss $\mathcal{L}_\mathrm{KLD}$ to the overall performance of our GMM-based method. Both losses prove to be effective and significantly enhance the source-only performance when used independently. Notably, the KL divergence loss thereby outperforms the contrastive loss, possibly because it impacts both the classifier and the feature extractor, whereas the contrastive loss only optimizes the feature extractor. However, the best results are achieved when combining both losses.

\begin{figure}[!t]
	\centering
	\includegraphics[width=0.9\linewidth]{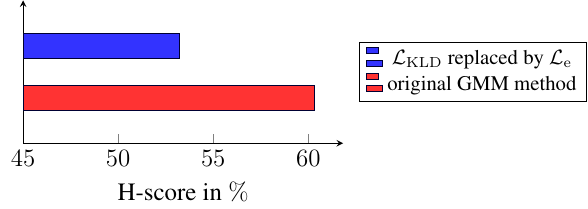}
	\caption{Results using COMET's entropy loss $\mathcal{L}_\mathrm{e}$ compared to the proposed KL divergence loss $\mathcal{L}_\mathrm{KLD}$ in our GMM method for the VisDA-C OPDA scenario.}
	\label{fig:entropy_loss}
\end{figure}

\paragraph{Comparison between the KL divergence loss and the entropy loss of COMET}
Our proposed KL divergence loss $\mathcal{L}_\mathrm{KLD}$ serves a similar function as the entropy loss $\mathcal{L}_\mathrm{e}$ used by COMET \cite{comet}. To evaluate their performance, we compare the original GMM method with a modified version where $\mathcal{L}_\mathrm{KLD}$ is replaced by COMET's entropy loss $\mathcal{L}_\mathrm{e}$. The results shown in \cref{fig:entropy_loss} indicate that using COMET's entropy loss instead of the KL divergence loss leads to a significant decline in performance. This suggests that the KL divergence loss is more effective at encouraging confident predictions for known class samples while also guiding the model to appropriately handle OOD samples, making it the better choice for SF-UniDA.


\subsection{Hyperparameter sensitivity}
\begin{figure}[!t]
	\centering
	\includegraphics[width=0.9\linewidth]{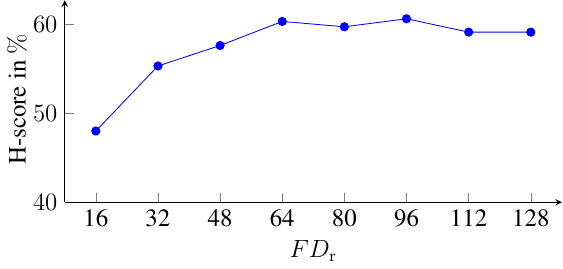}
	\caption{Results for the VisDA-C OPDA scenario using different numbers of dimensions for the reduced feature space.}
	\label{fig:red_feat}
\end{figure}

\begin{figure}[!t]
	\centering
	\includegraphics[width=0.9\linewidth]{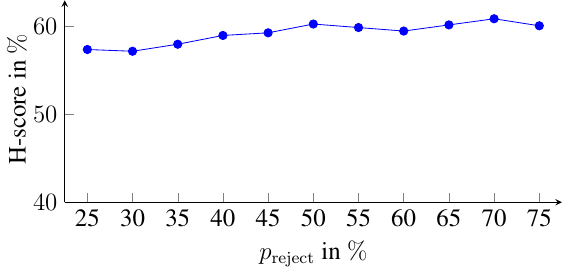}
	\caption{Results for the VisDA-C OPDA scenario using different rejection rates during pseudo-labeling.}
	\label{fig:p_reject}
\end{figure}

\begin{figure}[!t]
	\centering
	\includegraphics[width=0.9\linewidth]{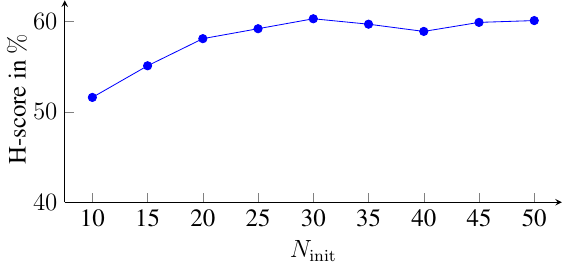}
	\caption{Results for the VisDA-C OPDA scenario using different numbers of batches for the initialization of $\tau_\mathrm{k}$ and $\tau_\mathrm{u}$ during pseudo-labeling.}
	\label{fig:N_init}
\end{figure}

\begin{figure}[!t]
	\centering
	\includegraphics[width=0.9\linewidth]{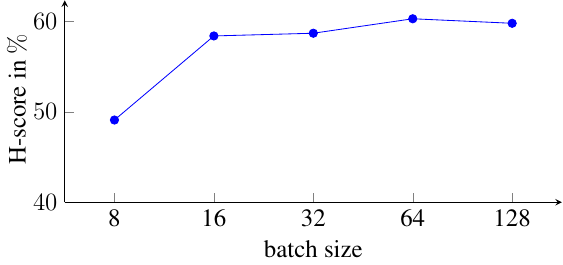}
	\caption{Results for the VisDA-C OPDA scenario using different batch sizes.}
	\label{fig:batch_size}
\end{figure}

\begin{figure}[!t]
	\centering
	\includegraphics[width=0.9\linewidth]{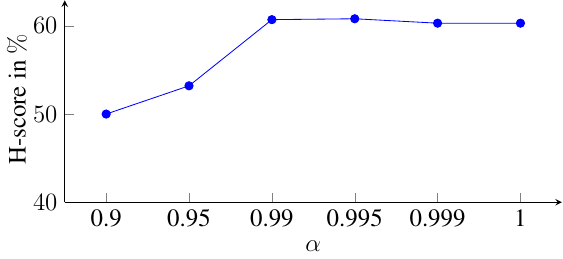}
	\caption{Results for the VisDA-C OPDA scenario using different exponential decay factors.}
	\label{fig:alpha}
\end{figure}

To analyze the sensitivity of our GMM-based method to hyperparameter choices, we vary each hyperparameter across a broad range around the chosen value while keeping the others constant. \cref{fig:red_feat,fig:p_reject,fig:N_init,fig:batch_size,fig:alpha} show the results. Overall, our approach proves to be robust against these variations which indicates that the choice of hyperparameters is not critical for its success.

\paragraph{Number of dimensions of the reduced feature space}
As shown in \cref{fig:red_feat}, stable performance is maintained when the number of dimensions $FD_\mathrm{r}$ of the reduced feature space is above $48$. There is minor degradation at $FD_\mathrm{r}=48$ and $FD_\mathrm{r}=32$, respectively, before the performance drops significantly at $FD_\mathrm{r}=16$. Therefore, the chosen number of feature dimensions $FD_\mathrm{r}=64$ seems to provide the best trade-off between performance and memory efficiency.

\paragraph{Rejection rate}
\cref{fig:p_reject} shows that the performance remains nearly constant when varying $p_\mathrm{reject}$ between $\SI{25}{\percent}$ and $\SI{75}{\percent}$. However, the performance slightly increases with a higher rejection rate. Therefore, discarding more samples during the initialization of $\tau_\mathrm{k}$ and $\tau_\mathrm{u}$ can be advantageous.


\paragraph{Number of initialization batches}
Similar to $FD_\mathrm{r}$ and $p_\mathrm{reject}$, the number of batches $N_\mathrm{init}$ used for initializing $\tau_\mathrm{k}$ and $\tau_\mathrm{u}$ does not significantly impact the performance of our method across a broad range of values, as shown in \cref{fig:N_init}. However, a performance degradation is observed for values smaller than $N_\mathrm{init}=20$. Thus, at least $20$ batches are necessary for a representative initialization of $\tau_\mathrm{k}$ and $\tau_\mathrm{u}$.

\paragraph{Batch size}
In \cref{fig:batch_size}, we observe that decreasing the batch size starting from $128$ only leads to a slight decline in performance until a batch size of $16$. 
However, a significant performance drop occurs at a batch size of $8$, which we suspect is due to the insufficient effectiveness of the initialization of $\tau_\mathrm{k}$ and $\tau_\mathrm{u}$ for such a small batch size. 

\paragraph{Exponential decay factor}
\cref{fig:alpha} shows that reducing the exponential decay factor from $\alpha=1$ (no decay) results in a small improvement until $\alpha=0.99$, after which performance declines sharply for $\alpha\leq0.95$. Therefore, selecting $\alpha>0.95$ appears to best balance the retention of valuable past information with the responsiveness to new data.

\begin{figure}[!t]
	\centering
	\includegraphics[width=0.9\linewidth]{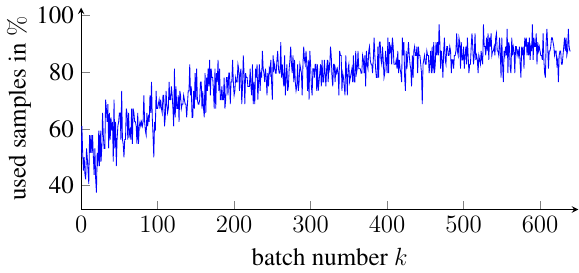}
	\caption{Development of the ratio of samples used for the adaptation over the course of an exemplary run of the VisDA-C OPDA scenario.}
	\label{fig:development}
\end{figure}

\subsection{Ratio of samples used for adaptation}
\cref{fig:development} illustrates how the percentage of samples used for adaptation evolves during the course of an exemplary run. For the first batch, $\SI{100}{\percent}-p_\mathrm{reject}=\SI{50}{\percent}$ of samples are used. During the initialization of $\tau_\mathrm{k}$ and $\tau_\mathrm{u}$ in the first $N_\mathrm{init}=30$ batches, this percentage remains relatively stable. Subsequently, it steadily increases on average, appearing to converge towards approx. $\SI{85}{\percent}$. This trend demonstrates the effectiveness of our adaptation method by showing that the pseudo-labeling becomes increasingly confident over time, leading to fewer samples being discarded due to uncertainty.

\subsection{OOD metric}
\begin{figure}[!t]
	\centering
	\includegraphics[width=0.9\linewidth]{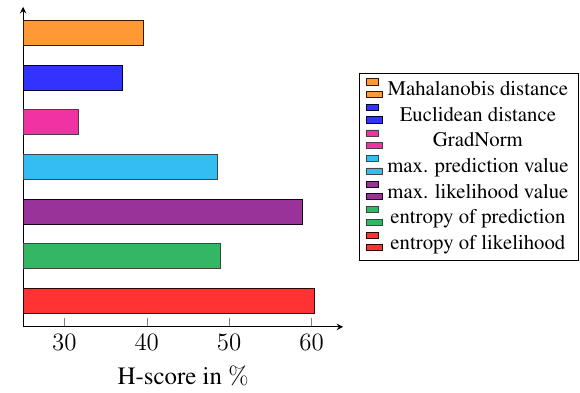}
	\caption{Results for the VisDA-C OPDA scenario using different metrics for the OOD detection.}
	\label{fig:ood_metric}
\end{figure}

Besides the entropy of the likelihood $I({\boldsymbol{p}}_{i,k})$, several other metrics can be used to evaluate the confidence of a sample belonging to the GMM distribution. These include the minimum Mahalanobis or Euclidean distance between the sample and the GMM means in the feature space \cite{mahalanobis}, GradNorm \cite{gradnorm}, the maximum prediction value $\max_{c\in\mathcal{Y}_\mathrm{s}} f_c(\boldsymbol{x}_{i,k}^\mathrm{t})$, the maximum likelihood value $\max_{c\in\mathcal{Y}_\mathrm{s}} p(\boldsymbol{x}_{i,k}^\mathrm{t} | c; \hat{\boldsymbol{\mu}}_{k}(c), \hat{\boldsymbol{\Sigma}}_{k}(c))$, and the entropy of the prediction $I(f(\boldsymbol{x}_{i,k}^\mathrm{t}))$. \cref{fig:ood_metric} presents a comparison of these OOD metrics for the VisDA-C OPDA scenario. It can be observed that the entropy of the likelihood provides the best result, while the maximum likelihood value performs only slightly worse. In contrast, all other evaluated OOD metrics yield significantly lower H-scores. Therefore, the two OOD metrics rooting on the likelihoods derived from the GMM prove to be the most robust, further demonstrating the effectiveness of our GMM-based knowledge transfer approach.

\subsection{Vision transformer as backbone architecture}
\cref{tab:ViT} presents the results on the VisDA-C dataset when the ResNet-50 backbone architecture is replaced with a Vision Transformer (ViT) \cite{vit}, specifically the ViT-B/16 model. Its source training follows the same procedure as that used for the ResNet-50. Qualitatively, the results are similar to those obtained with the ResNet-50 backbone reported in the main paper. Notably, our proposed GMM-based method continues to achieve the best performance across all category shifts and therefore proves to also be effective for ViT-based model architectures.

\begin{table}
	\caption{Results for the VisDA-C dataset using a Vision Transformer backbone. The accuracy (in $\%$) is reported for PDA, while the H-score (in $\%$) is reported for ODA and OPDA. Best results are in red, second best in blue.}
	\label{tab:ViT}
	\begin{tabularx}{\linewidth}{p{2.8cm} *{3}{c}}
		\toprule
		& PDA & ODA & OPDA\\
		\midrule
		Source-only & $19.5$ & $26.4$ & $19.4$\\
		\midrule
		OWTTT \cite{li2023robustness} & $31.3$ & $\color{blue}54.1$ & $\color{blue}49.9$\\
		COMET-P \cite{comet} & $\color{blue}36.7$ & $40.6$ & $38.8$\\
		\midrule
		SHOT-O/P \cite{shot} & $35.3$ & $48.0$ & $40.7$\\
		GLC \cite{sf_unida} & $9.9$ & $3.4$ & $9.6$\\
		GLC++ \cite{glc_plus} & $10.7$ & $5.3$ & $15.7$\\
		LEAD \cite{lead} & $4.2$ & $8.8$ & $8.8$\\
		COMET-F \cite{comet} & $31.3$ & $39.5$ & $38.6$\\
		GMM (Ours) & $\color{red}38.8$ & $\color{red}56.4$ & $\color{red}57.0$\\
		\bottomrule
	\end{tabularx}
\end{table}

\subsection{Measuring the absolute memory requirement}
To validate the theoretical memory consumption analysis presented in section 3.7 of the main paper and provide additional insights, we measure the absolute memory required for the knowledge transfer across batches of COMET \cite{comet}, a memory queue as proposed by \cite{AdaContrast}, and our GMM-based approach. Additionally, we measure and compare the overall peak memory usage of COMET and our GMM-based method during adaptation, which is crucial for determining hardware requirements.

First, we consider the VisDA-C OPDA scenario where the number of source classes is $|\mathcal{Y}_\mathrm{s}|=9$. COMET requires a copy of the ResNet-50-based model to implement the student-teacher architecture, which consumes $94,098.23\,\mathrm{KB}$ of memory. Second, in the memory queue, each prediction vector occupies $2.25\,\mathrm{KB}$, and each feature vector requires $8.00\,\mathrm{KB}$, resulting in a total of $10.25\,\mathrm{KB}$ per sample. Finally our GMM-based method requires $4.50\,\mathrm{KB}$ to store the means $\hat{\boldsymbol{\mu}}_{k}(c)$, $288.00\,\mathrm{KB}$ for the covariance matrices $\hat{\boldsymbol{\Sigma}}_{k}(c)$ and $0.07\,\mathrm{KB}$ for the weights $s_k(c)$, resulting in an overall memory consumption of $292.57\,\mathrm{KB}$. Therefore, our GMM-based approach only requires
\begin{align}
	\frac{292.57\,\mathrm{KB}}{94,098.23\,\mathrm{KB}}\approx0.0031=0.31\%
\end{align}
of memory compared to COMET. Moreover, starting with a memory queue size of only
\begin{align*}
	\left\lceil \frac{292.57\,\mathrm{KB}}{10.25\,\mathrm{KB}}\right\rceil=29
\end{align*}
the memory queue already consumes more memory than our proposed GMM-based method.

Next, we consider an arbitrary OPDA scenario of DomainNet where the number of source classes is $|\mathcal{Y}_\mathrm{s}|=200$. Due to the increased number of output neurons, the memory consumption of the ResNet-50-based model slightly increases to $94,290.73\,\mathrm{KB}$. Regarding the memory queue, each prediction vector now occupies $50.00\,\mathrm{KB}$ resulting in a total of $58.00\,\mathrm{KB}$ per sample. For the GMM-based method, the memory requirements increase to $100.00\,\mathrm{KB}$ for the means, $6,400.00\,\mathrm{KB}$ for the covariance matrices, and $1.56\,\mathrm{KB}$ for the weights, totaling $6,501.56\,\mathrm{KB}$. Hence, in this case, our GMM-based approach uses only
\begin{align}
	\frac{6,501.56\,\mathrm{KB}}{94,290.73\,\mathrm{KB}}\approx0.0690=6.90\%
\end{align}
of the memory compared to COMET. Moreover, now starting from a queue size of
\begin{align*}
	\left\lceil \frac{6,501.56\,\mathrm{KB}}{58.00\,\mathrm{KB}}\right\rceil=113
\end{align*}
the memory queue exceeds the memory usage of our GMM-based method. Although this value is considerably higher than for the VisDA-C dataset, it corresponds to significantly fewer than one sample per source class, which is still insufficient to enable reliable kNN- or clustering-based pseudo-labeling.

A similar picture is also observed when measuring the overall peak memory usage of COMET and our GMM-based method during each adaptation step. For the VisDA-C OPDA scenario, COMET's peak memory usage is nearly three times higher at $16,115.05,\mathrm{MB}$ compared to just $5,792.14,\mathrm{MB}$ for our GMM-based approach. Similarly, in the DomainNet OPDA scenario, COMET requires more than twice the memory, with a maximum of $16,113.88,\mathrm{MB}$ compared to $6,954.79,\mathrm{MB}$ for our method. Thus, as expected, the overall memory footprint of our GMM-based method is significantly smaller than that of COMET due to its more memory-efficient knowledge transfer. This reduced memory usage lowers hardware requirements, making our approach more suitable for real-world applications, particularly in embedded systems.

%
%
%

\section{Discussions}
\subsection{Potential weaknesses of our method}
Our method is based on the assumption that the target data classes form unimodal Gaussian-distributed clusters in the feature space. However, this may not always hold true. Although the contrastive loss encourages this clustering behavior, if the pseudo-labels are inaccurate at the beginning of the adaptation process due to a violation of this assumption, it can result in a slow initial adaptation. In the worst case, if the pseudo-labels are too inaccurate to trigger effective clustering, the adaptation may even fail completely. Although we did not encounter such a case in our experiments, to mitigate this risk, we recommend opting for a higher rather than lower initial ratio of left out samples $p_\mathrm{reject}$.

A second potential failure case could occur with a highly imbalanced target dataset that contains only few samples of unknown classes. In this situation, since our KL divergence loss maximizes the divergence between the classifier output and a uniform distribution for samples pseudo-labeled as a known class, it may cause the classifier to converge to a trivial solution, i.e. always predicting the same class with high confidence. To prevent this from happening, in such a case, it might be advantageous to instead minimize the KL divergence between the prediction vector and the one-hot encoded pseudo-label for samples pseudo-labeled as a known class. By doing so, and maintaining a uniform pseudo-label for the unknown class, the KL divergence loss effectively becomes a cross-entropy loss. Nevertheless, also this case has never occurred during our experiments.

\subsubsection{Difference between online and offline setting}
When comparing the results from our online scenario with those achieved in offline SF-UniDA, like the results provided by GLC \cite{sf_unida}, GLC++ \cite{glc_plus}, and LEAD \cite{lead}, a notable difference is evident. In the offline scenario significantly higher scores are achieved compared to online. For instance, the performance gap in the DomainNet OPDA scenario (considering only the domains painting, real, and sketch) is around $\SI{5}{\percent}$, while it extends to approximately $\SI{15}{\percent}$ in the VisDA-C OPDA scenario.

We believe that two main factors contribute to this discrepancy. First, offline methods have the advantage of being able to access all target data at once, which allows for more thorough adaptation strategies, such as kNN, which are not feasible in the online setting. Second, in offline scenarios, adaptation and prediction are separate steps, meaning the prediction only starts once the adaptation has been finished. This leads to an improved performance right from the start of the prediction. In contrast, in the online setting prediction and adaptation need to be performed in parallel which means that the prediction performance is initially equal to the source-only performance and only improves gradually. This difference in the adaptation process naturally reflects on the average performance scores.

\end{document}